\documentclass[conference]{IEEEtran}
\IEEEoverridecommandlockouts
\usepackage{cite}
\usepackage{amsmath,amssymb,amsfonts}
\usepackage{graphicx}
\usepackage{textcomp}
\usepackage{xcolor}
\def\BibTeX{{\rm B\kern-.05em{\sc i\kern-.025em b}\kern-.08em
    T\kern-.1667em\lower.7ex\hbox{E}\kern-.125emX}}

\usepackage[colorlinks,urlcolor=blue,linkcolor=blue,citecolor=blue]{hyperref}

\usepackage{color,array}

\usepackage{graphicx} 
\usepackage{float}
\usepackage{subcaption}

\usepackage{algorithm}
\usepackage{algpseudocode}

\usepackage{amsmath}
\DeclareMathAlphabet\mathbfcal{OMS}{cmsy}{b}{n}
\usepackage{bm}

\usepackage{lipsum}

\usepackage{stfloats}
\usepackage{bbold}

\begin{document}

\title{Adversarial Detection: Attacking Object Detection in Real Time\\
\thanks{The project is supported by Offshore Robotics for Certification of Assets (ORCA) Partnership Resource Fund (PRF) on Towards the Accountable and Explainable Learning-enabled Autonomous Robotic Systems (AELARS) [EP/R026173/1].
.}
}

\makeatletter
\newcommand{\linebreakand}{%
  \end{@IEEEauthorhalign}
  \hfill\mbox{}\par
  \mbox{}\hfill\begin{@IEEEauthorhalign}
}
\makeatother

\author{
\IEEEauthorblockN{1\textsuperscript{st} Han Wu}
\IEEEauthorblockA{\textit{Computer Science} \\
\textit{The University of Exeter}\\
Exeter, the United Kingdom \\
hw630@exeter.ac.uk}
\and
\IEEEauthorblockN{2\textsuperscript{nd} Syed Yunas}
\IEEEauthorblockA{\textit{Computer Science} \\
\textit{The University of the West of England}\\
Bristol, the United Kingdom \\
syed.yunas@uwe.ac.uk}
\and
\IEEEauthorblockN{3\textsuperscript{rd} Sareh Rowlands}
\IEEEauthorblockA{\textit{Computer Science} \\
\textit{The University of Exeter}\\
Exeter, the United Kingdom \\
s.rowlands@exeter.ac.uk} \\
\linebreakand
\IEEEauthorblockN{4\textsuperscript{th} Wenjie Ruan}
\IEEEauthorblockA{\textit{Computer Science} \\
\textit{The University of Exeter}\\
Exeter, the United Kingdom \\
w.ruan@exeter.ac.uk}
\and
\IEEEauthorblockN{5\textsuperscript{th} Johan Wahlstr\"om$^{*}$}
\IEEEauthorblockA{\textit{Computer Science} \\
\textit{The University of Exeter}\\
Exeter, the United Kingdom \\
j.wahlstrom@exeter.ac.uk}
}

\maketitle

\begin{abstract}

Intelligent robots rely on object detection models to perceive the environment. Following advances in deep learning security it has been revealed that object detection models are vulnerable to adversarial attacks. However, prior research primarily focuses on attacking static images or offline videos. Therefore, it is still unclear if such attacks could jeopardize real-world robotic applications in dynamic environments. 
This paper bridges this gap by presenting the first real-time online attack against object detection models. We devise three attacks that fabricate bounding boxes for nonexistent objects at desired locations. The attacks achieve a success rate of about 90\% within about 20 iterations. The demo video is available at \href{https://youtu.be/zJZ1aNlXsMU}{https://youtu.be/zJZ1aNlXsMU}.


\end{abstract}


\section{Introduction}

Reliable object detection is crucial for multiple safety-critical robotic applications. For example, an autonomous driving system relies on object detection models to perceive the environment and take action on it. While advances in deep neural networks have rapidly increased the availability of high-accuracy object detection models, this breakthrough has also revealed several potential vulnerabilities. The first adversarial attacks fooled image classification models by adding imperceptible perturbations to the input image \cite{goodfellow2014explaining}. Later research removed the restriction to imperceptible perturbations and instead designed an adversarial patch that can be printed in the physical world \cite{brown2017patch}. The physical patch fooled an image classification model to misclassify the most salient object in the image. However, attacks against image classification models cannot fool object detection models. Therefore, in 2018, Liu et al. introduced the DPatch, a digital patch that fools object detection models \cite{liu2018dpatch}, and Lee et al. later extended the attack to physical patches \cite{lee2019physical}.

The generation of adversarial patches for attacking real-time robotic applications is still a rather challenging task. The generation process itself is often very computationally expensive. In addition, the efficiency of the physical patch is conditioned on strict requirements on the relative distance and orientation of the camera and the patch \cite{wang2021daedalus} \cite{9762099}. These requirements are often difficult to satisfy for real-world robotic applications deployed in dynamic environments.

\begin{figure}[tbp]
    \centering
    \begin{subfigure}[b]{0.48\textwidth}
        \includegraphics[width=\linewidth]{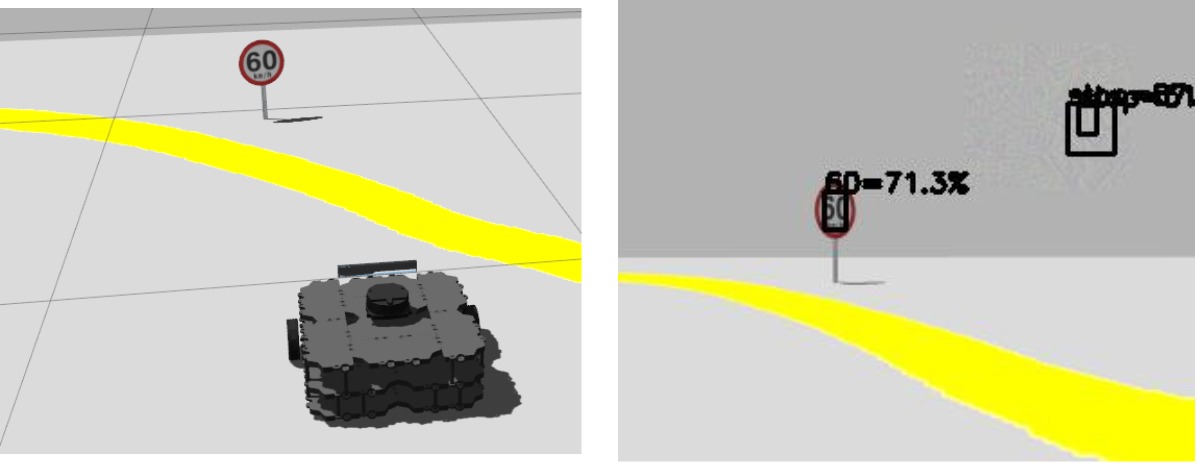}
        \caption{The \textbf{original image} (left) only contains one traffic sign. The \textbf{One-Targeted} Attack (right) generates one overlay containing several target objects (stop sign).}
        \label{fig:sparse-untar} 
    \end{subfigure}

    \begin{subfigure}[b]{0.48\textwidth}
        \includegraphics[width=1\linewidth]{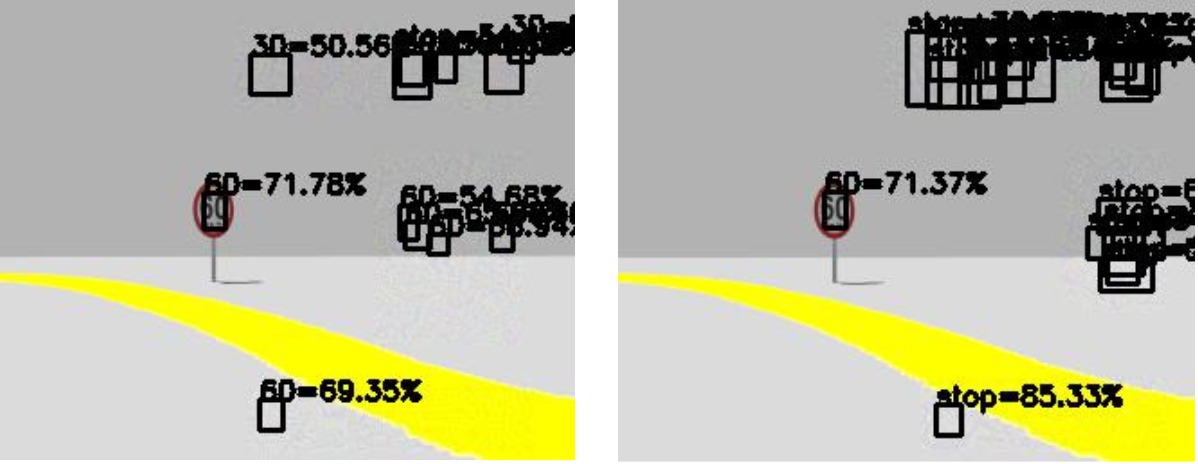}
        \caption{The \textbf{Multi-Untargeted} Attack (left) generates multiple overlays containing different kinds of objects (30, 60, stop). The \textbf{Multi-Targeted} Attack (right) generates multiple overlays containing several target objects (stop sign).}
        \label{fig:dense-untar}
    \end{subfigure}
    \caption{This paper demonstrates how to generate adversarial overlays of arbitrary shapes at specified positions in real time.}
    \label{fig:overview}
\end{figure}

Recent research in penetration tests against the Robot Operating System (ROS) has shown that it is possible to inject digital patches into the ROS message that contains the camera image \cite{dieber2020penetration}. In particular, Dieber et al. succeeded in isolating a ROS node so that they could intercept and manipulate the ROS message that contains sensor data while the victim node is unaware of the attack. In this paper, we generate digital overlays (unperceivable digital patches) at desired locations in real time and then exploit the vulnerability to inject the digital overlays into the input image. Overall, this paper makes the following contributions:
\begin{itemize}
    \item We devise three adversarial attacks that can generate digital overlays of different shapes at specified locations in real time (see Fig. \ref{fig:overview}). This extends previous research that primarily has focused on square patches.
    \item 
    Investigating the effect of various digital overlays it is found that the size of the overlay is critical for the 
    success of the attack, whereas the attack performance is independent of the aspect ratio.
    
    \item We test our attacks in the Robot Operating System (ROS). The system is open-sourced to faciliate future extensions and comparisons\footnote{The code is available on Github: \url{https://github.com/wuhanstudio/adversarial-detection}}.
\end{itemize}


\section{Preliminaries}
\label{section_preliminaries}

This section clarifies the differences between adversarial filters and adversarial patches. Following this, we describe how prior research applies these filters and patches in the digital and physical world. \emph{Our} research focuses on applying adversarial patches in the digital world.

\begin{figure}[b]
    \centering
    \begin{subfigure}[b]{0.48\textwidth}
        \includegraphics[width=1\linewidth]{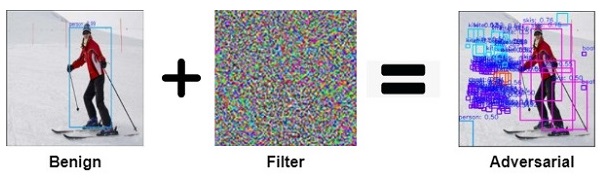}
        \caption{Digital filters apply the perturbation to the entire input image in the digital world \cite{wang2021daedalus}.}
        \label{fig:digital_filter}
    \end{subfigure}
    \begin{subfigure}[b]{0.48\textwidth}
        \includegraphics[width=1\linewidth]{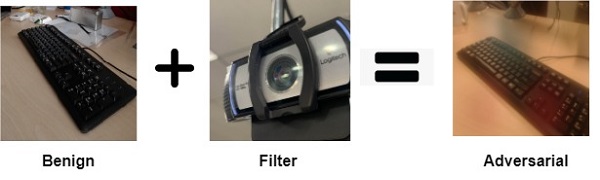}
        \caption{Physical filters apply the perturbation by attacking the camera using, e.g., a translucent film attached to the camera lens \cite{li2019adversarial}.}
        \label{fig:physical_filter}
    \end{subfigure}
    \caption{Adversarial filters perturb the entire image.}
    \label{fig:filter}
\end{figure}

\subsection{The Adversarial Filter}

An adversarial filter refers to a perturbation added to the entire image. Thus, the adversarial filter is of the same size as the input image and is unperceivable by human eyes. Based on where we apply the perturbation, adversarial filters can be categorized as digital or physical filters.

\subsubsection{\textbf{Digital Filter} (Fig. \ref{fig:digital_filter})}

The first adversarial attack against image classification models was a digital filter that added a small perturbation to the entire input image \cite{goodfellow2014explaining}. The perturbation can be generated using either gradient-based methods \cite{madryMSTV18} \cite{kurakin2018adversarial} \cite{wong2019wasserstein} \cite{croce2020reliable} or optimization-based methods \cite{papernot2016transferability} \cite{carlini2017towards} \cite{qin2019imperceptible}. Prior research uses the $l_1$, $l_2$, and $l_{\infty}$ norms to measure the human perceptual distance between the adversarial and the original input image \cite{miyato2015distributional} \cite{sabour2015adversarial} \cite{chen2018ead}. While the digital filter first proved the existence of adversarial examples, it is limited in practical scenarios since it is not always possible to assume access to the input image.

\subsubsection{\textbf{Physical Filter} (Fig. \ref{fig:physical_filter})}

Physical filters attach a translucent film to the camera lens to perturb the entire image \cite{li2019adversarial}. While physical filters do not require access to the input image, they do require physical access to the camera. One big challenge with physical filters is the difficulty of manufacturing a film that precisely replicates the adversarial perturbation. Thus, the practical applicability of adversarial filters in the physical world is still rather limited. In addition, to the best of our knowledge, physical filters has so far only been used to attack image classification models, not object detection models. 


\begin{figure}[b]
    \centering
    \begin{subfigure}[b]{0.48\textwidth}
        \includegraphics[width=1\linewidth]{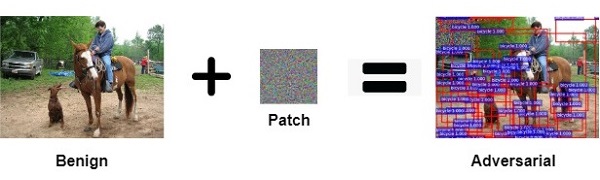}
        \caption{Digital patches replace a part of the input image with the adversarial patch (left top corner) \cite{liu2018dpatch}.}
        \label{fig:digital_patch}
    \end{subfigure}
    \begin{subfigure}[b]{0.48\textwidth}
        \includegraphics[width=1\linewidth]{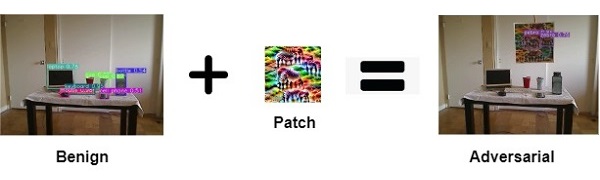}
        \caption{Physical patches print the adversarial patch on a physical object, e.g., on a poster \cite{lee2019physical}.}
        \label{fig:physical_patch}
    \end{subfigure}
    \caption{Adversarial patches perturb part of the image.}
    \label{fig:patch}
\end{figure}

\subsection{The Adversarial Patch}

In some scenarios, it is possible to remove the constraint of the imperceptibility of the perturbation. This means that we can generate so-called patches that are clearly distinguishable to anyone viewing the image in question. 

Contrary to filters, these patches are limited in size and only perturb part of the image, allowing us to better control the location of generated objects when attacking an object detection model. Similar to filters, patches can be categorized as digital or physical patches.

\subsubsection{\textbf{Digital Patch} (Fig. \ref{fig:digital_patch})} In 2018, Liu etc al. designed the DPatch \cite{liu2018dpatch} to attack the Faster R-CNN object detection model \cite{ren2015faster}. The DPatch is location-independent, which means that it can be placed anywhere in the input image. Another notable digital patch is the TOG attack, presented in 2020 by Chow et al \cite{chow2020adversarial}. They generated a 40x40 adversarial patch to fool object detection models into misclassifying objects to which the patch was attached.

\subsubsection{\textbf{Physical Patch} (Fig. \ref{fig:physical_patch})} In 2017, Google presented landmark research for generating a physical patch to attack an image classification model \cite{brown2017patch}. As a result, research interest gradually shifted to from digital to physical patches. Later, Lee et al. \cite{lee2019physical} and Wang et al. \cite{wang2021daedalus} extended the physical patch attack from image classification to object detection. One shortcoming of physical patches is their inflexibility. Once the adversarial patch has been printed out, the only way to change it is through reprinting. To improve their flexibility, Hoory et al. dynamically displayed adversarial patches on a flat screen attached to a car \cite{hoory2020dynamic}.




\section{Adversarial Detection}

Our objective is to fool the YOLOv3 \cite{redmon2018yolov3}, and YOLOv4 \cite{bochkovskiy2020yolov4} object detection models to mistakenly detect objects at locations where there is no object. We generate adversarial overlays by combining the imperceptibility of adversarial filters with the localizability of adversarial patches.

\subsection{Problem Formulation} 

The YOLO object detection model splits the input image into $S$x$S$ grids and makes predictions for each grid. For example, the input shape of YOLO is 416x416x3 (height, width, and channel). If $S=13$, we divide each channel into 13x13 grids, and each grid is 32x32 pixels. To detect objects of different sizes, YOLO makes predictions at three different scales (see Fig. \ref{fig.grid}). The first, second, and third output layer contains 13x13, 26x26, and 52x52 grids, respectively. The first output layer detects larger objects, and the third output layer detects smaller objects. In addition, YOLO pre-defines three anchor boxes ($B=3$) at each scale to detect objects of different aspect ratios. Thus, we have 9 pre-defined anchor boxes for 3 scales. Lastly, each output contains the shape and location of the bounding box, the confidence value, and the probability vector for each class (see Fig. \ref{fig.grid}). For example, if the model is pretrained on the MS COCO dataset \cite{mscoco2014} \cite{moore2020fiftyone} that contains 80 classes ($K=80$), each output contains 85 values consisting of four dimensions ($b_x, b_y, b_w, b_h$), one confidence value ($c$), and 80 probabilities $(p_1, p_2, ..., p_{80})$ for each class.

Putting things together: Given an input image $x$, the object detection model outputs $S$x$S$ candidate bounding boxes $o \in \mathcal{O}$ at three different scales ($S \in \{13,26,52\}$, $B=3$, and $|\mathcal{O}| = \sum_{1 \leq i \leq 3} S_i \times S_i \times B$, where $S_i$ represents the grid size of the $i_{th}$ output layer). Further, each candidate box is
\begin{equation}
 o^i = [b_x^i, b_y^i, b_w^i, b_h^i, c^i, p_1^i, p_2^i, ..., p_K^i],\ 1 \leq i \leq |\mathcal{O}|
\end{equation}
for $K$ classes. For example, the shape of the first output layer is $(S_1,\ S_1,\ B,\ K + 5) = (13,\ 13,\ 3,\ 85)$. The image is divided into 13x13 grids, and the model makes predictions for 3 anchor boxes. Each prediction contains 85 values (85 = 4 + 1 + 80). The adversarial attack aims to generate an adversarial perturbation $\delta \in [-1, 1]^{whc}$ such that the adversarial output $\mathcal{O}(x^{'})$ is different from the original output $\mathcal{O}(x)$.

\begin{figure}[b]
    \centering
    \includegraphics[width=0.5\textwidth]{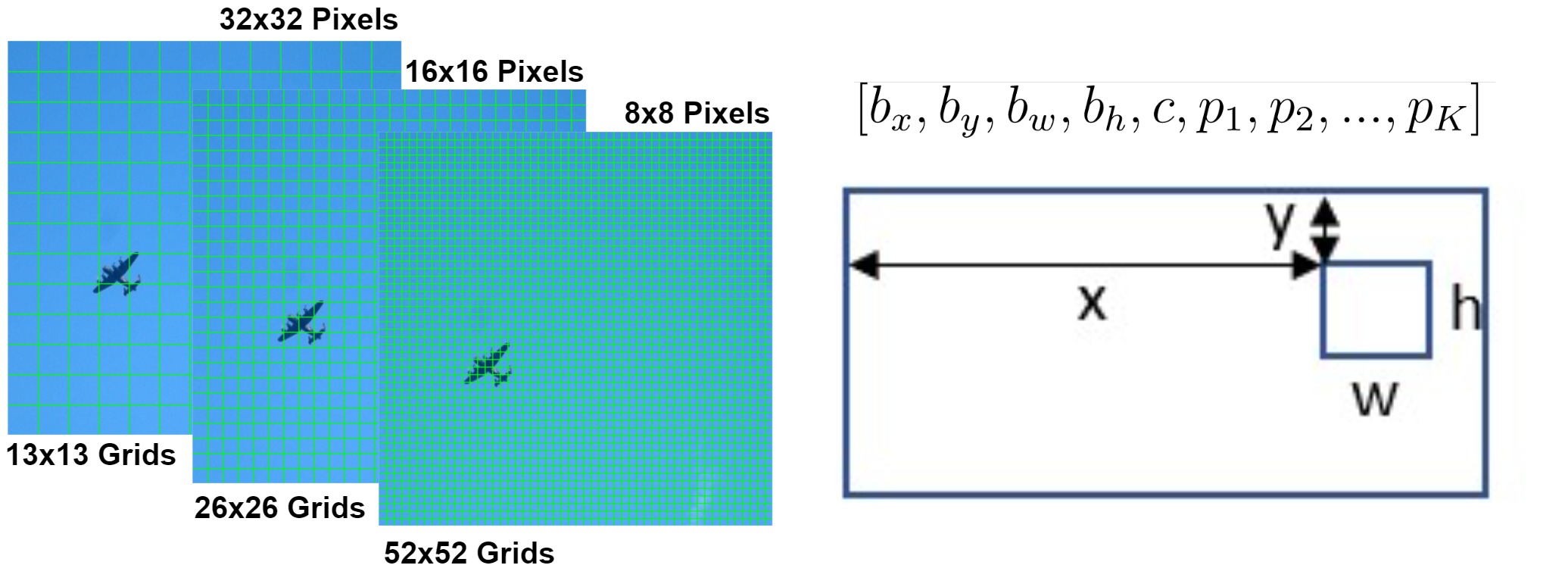}
    \caption{The YOLO object detection model makes predictions at three scales.}
    \label{fig.grid}
\end{figure}


\subsection{The Adversarial Overlay}

As described in Section \ref{section_preliminaries}, adversarial filters are imperceptible to human eyes but do not allow us to control exactly where in an image that we fabricate objects. Adversarial patches, 
on the other hand, are pinpointable but conspicuous. We propose a new method, the adversarial overlay, that applies an imperceptible perturbation to part of an image. Thereby, the adversarial overlay combines the strengths of adversarial filters and patches. The perturbations applied in filters, patches, and overlays can be contrasted as 
\begin{subequations}
\label{eq_perturbations}
\begin{align}
    x^{'}_{filter} &= x + \delta, \\
    x^{'}_{patch} &= (\mathbb{1}-m) \odot x + m \odot \delta, \\
    x^{'}_{overlay} &= x + m \odot \delta.
\end{align}
\end{subequations}
Here, the perturbation $\delta$ is applied via a binary mask $m \in \{0, 1\}^{wh}$, and $\odot$ is the Hadamard operator performing element-wise multiplication. As seen in equation \eqref{eq_perturbations}, the adversarial filter $x^{'}_{filter}$ applies the perturbation to the entire image without the mask $m$, the adversarial patch $x^{'}_{patch}$ uses the mask to replace part of the image, and the adversarial overlay $x^{'}_{overlay}$ applies the mask to the perturbation and adds it as an overlay to the image.

Now, note that the confidence value and the probability vector collectively determine whether or not a bounding box is drawn on the output image. Thus, by maximizing the product of confidence and probability vector, we can fabricate objects at desired locations. This leads us to propose three adversarial loss functions $L_{adv}$ for generating perturbations in real time: one loss function for one-targeted attacks
\setcounter{equation}{2}
\begin{equation}
\max_{1 \leq i \leq |\mathcal{O}|}\ [\sigma(c^i) * \sigma(p^i_t)] \label{eq:one-targeted},
\end{equation}
one loss function for multi-targeted attacks 
\begin{equation}
\sum^{|\mathcal{O}|}_{i = 1}\ [\sigma(c^i) * \sigma(p^i_t)] \label{eq:multi-targeted},    
\end{equation}
and one loss function for multi-untargeted attack
\begin{equation}
\sum^{|\mathcal{O}|}_{i = 1} \sum_{j=1}^{K}\ [\sigma(c^i) *\sigma(p^i_j)] \label{eq:multi-untargeted}.    
\end{equation}
Here, the function $\sigma (x)$ represents the sigmoid function.




\textbf{The one-targeted attack} (Eq. \ref{eq:one-targeted}) generates one overlay that contains the target object $t \in \{1,\dots,K\}$ by finding the bounding box with the maximum value of $\sigma(c^i) * \sigma(p^i_t)$. This method generates only one overlay that contains target objects because we increase the confidence and probability of the maximum one in all candidate bounding boxes.

\textbf{The multi-targeted attack} (Eq. \ref{eq:multi-targeted}) generates multiple overlays that contain target objects $t \in [0, K]$ by maximizing the sum of $\sigma(c^i) * \sigma(p^i_t)$ for the target class.

\textbf{The multi-untargeted attack} (Eq. \ref{eq:multi-untargeted}) generates multiple overlays that contain different objects by maximizing the sum of $\sigma(c^i) * \sigma(p^i_t)$ for all classes.

Note that it is also possible to generate monochrome greyscale overlays. Greyscale overlays add the same value to each channel, making the perturbation less conspicuous. We can generate monochrome overlays using the average of the red, green, and blue channels or from a single channel since human eyes are most sensitive to the green channel.

The perturbation is computed using gradients. At the end of each iteration, we clip the value of the perturbation so that it does not exceed the pre-defined boundary. The adversarial image of the original DPatch \cite{liu2018dpatch} contains invalid negative pixel values. Thus, we also clip the value of the adversarial image to make sure it is still a valid image. The adversarial overlay attack is summarized in Algorithm \ref{alg:adv-overlay}.

\begin{figure*}[b]
\centering
\begin{subfigure}[b]{0.31\textwidth}
    \centering
    \includegraphics[width=\textwidth]{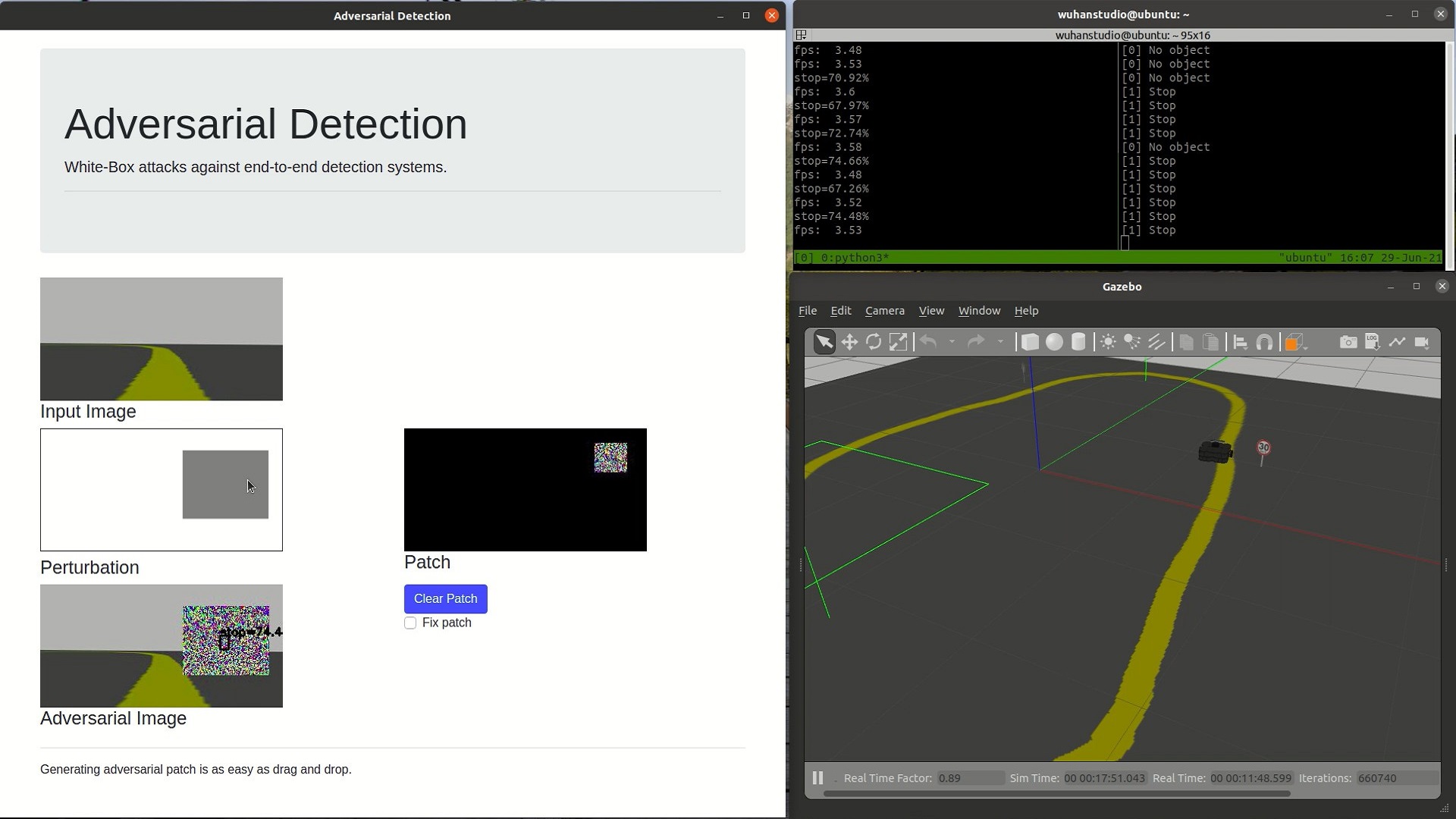}
    \caption{The Adversarial Overlay in Gazebo.}
    \label{fig:gazebo}
\end{subfigure}
\hfill
\begin{subfigure}[b]{0.31\textwidth}
    \centering
    \includegraphics[width=\textwidth]{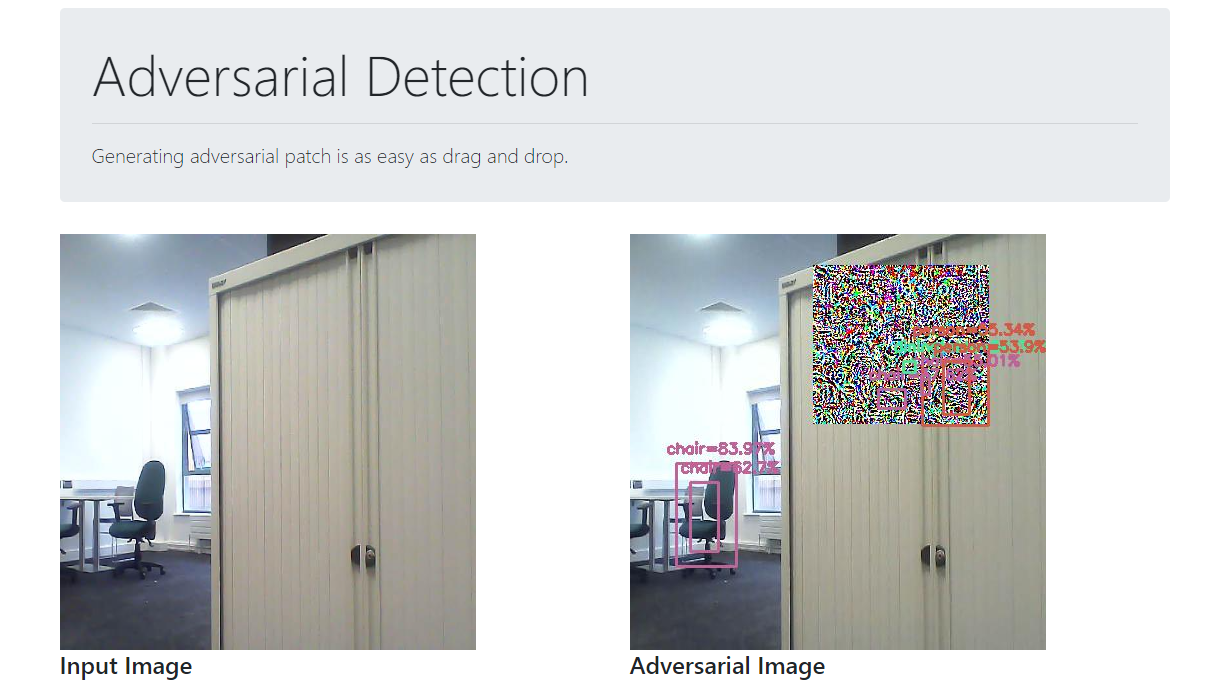}
    \caption{The Adversarial Overlay on a Laptop.}
    \label{fig:pc}
\end{subfigure}
\hfill
\begin{subfigure}[b]{0.31\textwidth}
    \centering
    \includegraphics[width=\textwidth]{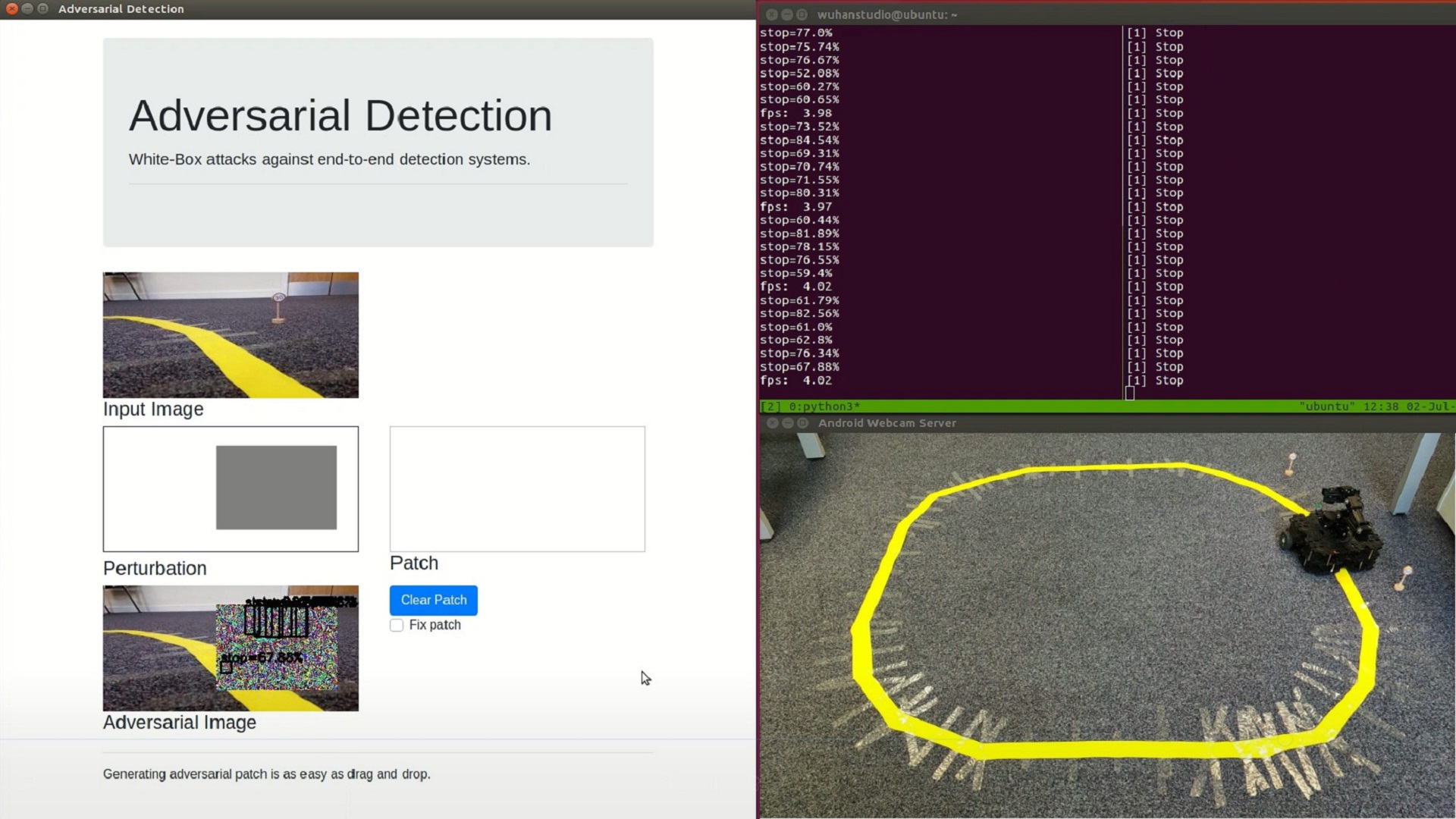}
    \caption{The Adversarial Overlay on Turtlebot3.}
    \label{fig:turtlebot}
\end{subfigure}
\caption{We tested our attack in three environments. Here we made the overlay visible for illustration purposes.}
\label{fig.adv_detect_demo}
\end{figure*}

\begin{algorithm}[t]
    \caption{The Adversarial Overlay Attack}\label{alg:adv-overlay}
    \begin{algorithmic}
        \State Input: The object detection model $f(\theta, x)$, a mask $m$.
        \State Parameters: The number of iterations $n$, the learning rate $\alpha$, and the strength of the attack $\xi$ measured by $l_{\infty}$ norm.
        \State Output: The adversarial perturbation $\delta$.
        \State \textbf{Initialization}: 
        \If {monochrome}
            \State $\delta \leftarrow 0^{416\text{x}416}$
        \Else
            \State $\delta \leftarrow 0^{416\text{x}416\text{x}3}$
        \EndIf
        \For{each input image $x$}
            \State $x' = x$
            \For{each iteration}
                \If {monochrome}
                \State R Channel: $x_R^{'} = x_R' + m \odot \delta$
                \State G Channel: $x_G^{'} = x_G' + m \odot \delta$
                \State B Channel: $x_B^{'} = x_B' + m \odot \delta$
                \Else
                    \State Overlay: $x^{'} = x' + m \odot \delta$
                \EndIf
                \State Gradient: $\nabla = \frac{\partial \mathcal{L}_{adv}(\mathcal{O})}{\partial x'}$
                \If {monochrome}
                    \State $\delta = \delta +  \frac{1}{3}\alpha(\nabla_R + \nabla_G + \nabla_B) $
                \Else
                    \State $\delta = \delta + \alpha \cdot \text{sign}(\nabla)$
                \EndIf
                \State $\delta = \text{clip}(\delta,\ -\xi,\ \xi)$
                \State $x^{'} = \text{clip}(x^{'},\ 0.0,\ 1.0)$
            \EndFor
        \EndFor
    \end{algorithmic}
\end{algorithm}



\subsection{System Architecture}

This papers presents an adversarial detection system to attack the YOLO object detection model. The system adopts a modular design pattern so that we can test our attacks in different environments (see Fig. \ref{fig.adv_detect_demo}) using the same architecture. The system consists of three components: the Data Source, the Server, and the Control Panel.

\textbf{The Data Source}: The data source publishes the input image to the server. We can publish the image from different sources, including a PC camera, the ROS Gazebo Simulator, and a real Turtlebot 3.

\textbf{The Server}: The server receives the input image stream from the data source via WebSocket connections. Meanwhile, it obtains the adversarial mask from the control plane. The server then generates and injects the adversarial overlay into the input image. In addition, the YOLO object detection model is deployed on the server.

\textbf{The Control Plane}: The control plane is a website where the attacker can draw the mask at arbitrary locations. Then, the browser sends the mask to the server via Websocket connections.


\clearpage

\section{Experimental Results}

We used YOLO object detection models pre-trained on the MS COCO dataset for the PC environment. Moreover, we trained two traffic sign detection models using YOLO for the ROS Gazebo and the ROS Turtlebot environment. 

\begin{figure*}[t]
\centering
\begin{subfigure}[b]{0.32\textwidth}
    \centering
    \includegraphics[width=\textwidth]{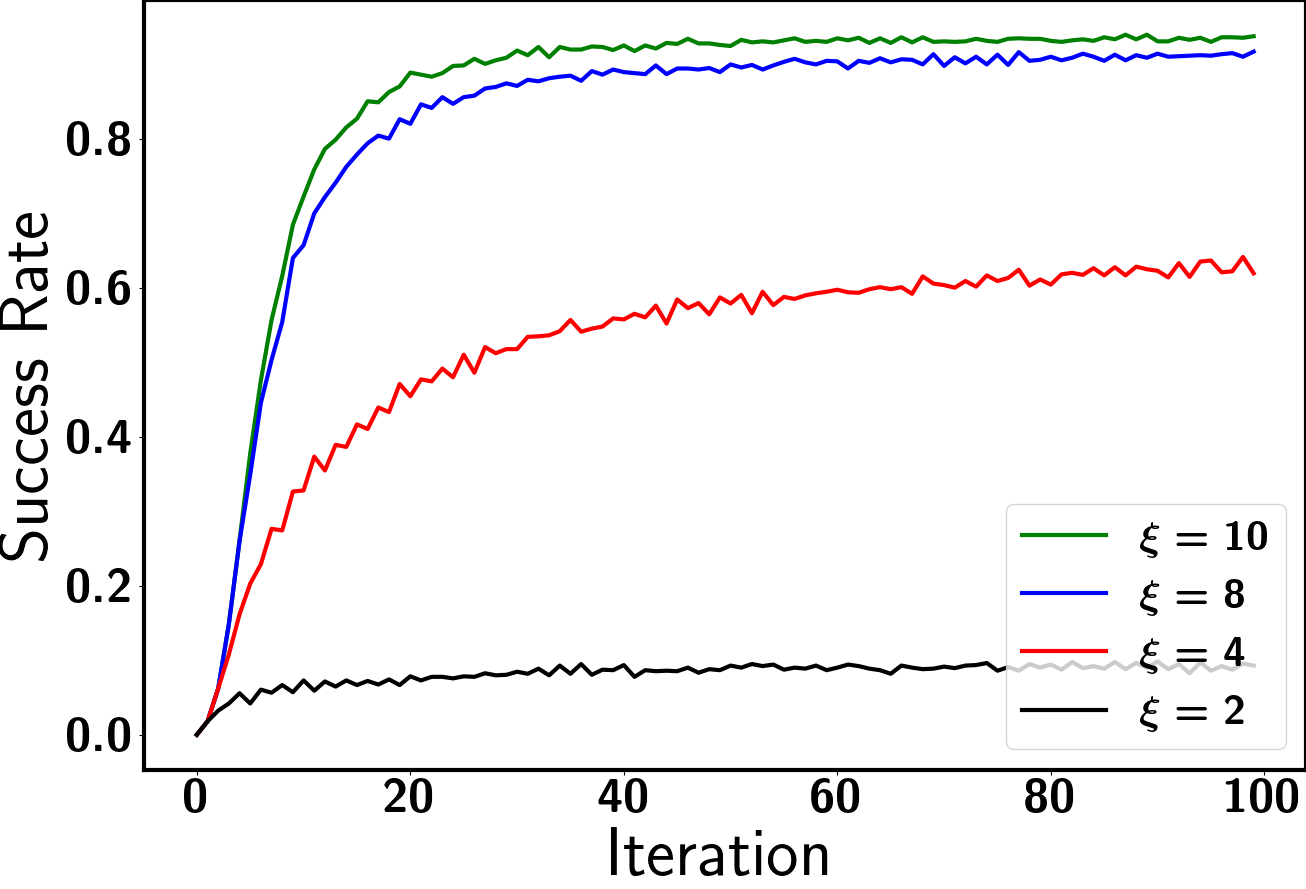}
    \caption{Different $\xi$ with $\alpha=2$, box size$=64$.}
    \label{fig:xi_suc}
\end{subfigure}
\hfill
\begin{subfigure}[b]{0.32\textwidth}
    \centering
    \includegraphics[width=\textwidth]{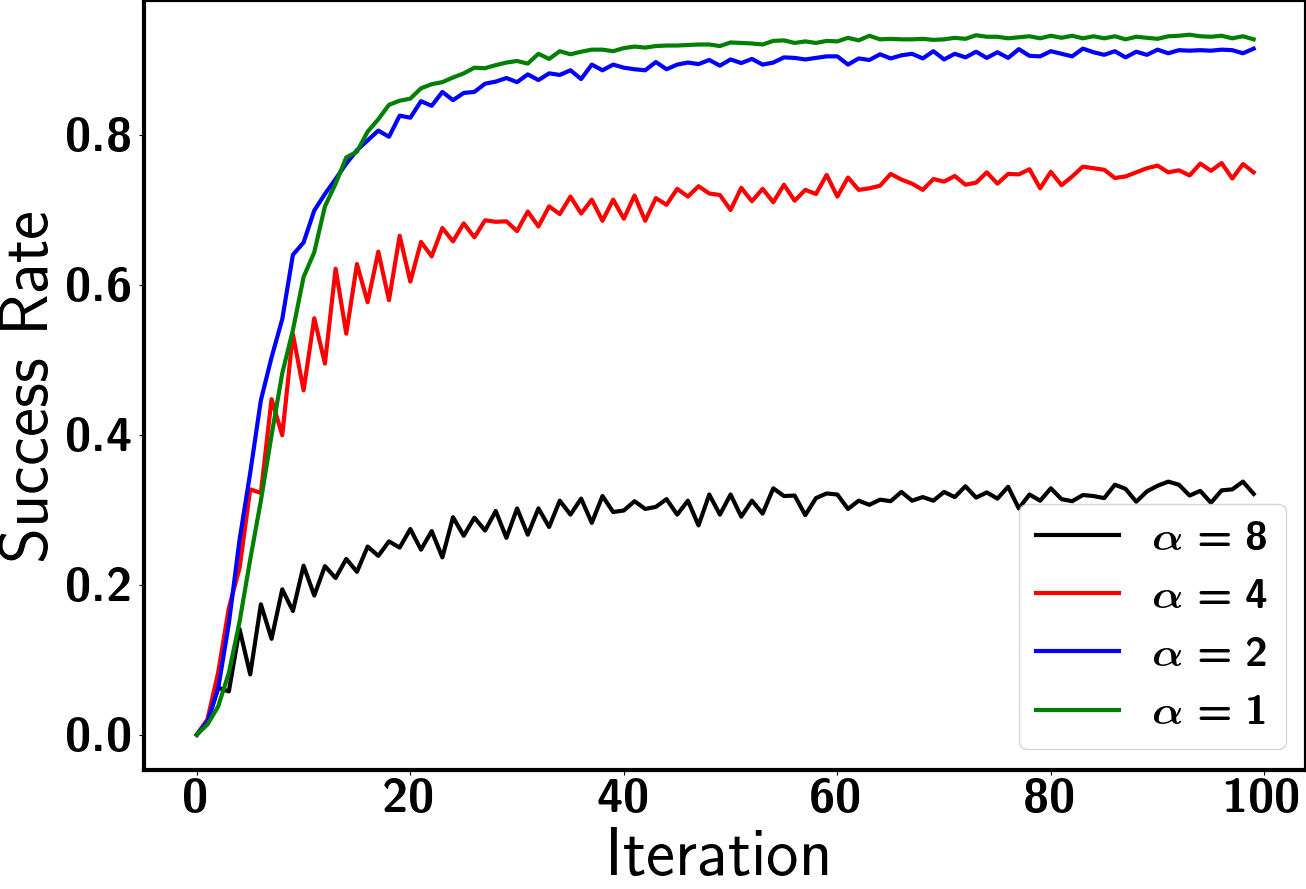}
    \caption{Different $\alpha$ with $\xi=8$, box size $=64$.}
    \label{fig:alpha_suc}
\end{subfigure}
\hfill
\begin{subfigure}[b]{0.32\textwidth}
    \centering
    \includegraphics[width=\textwidth]{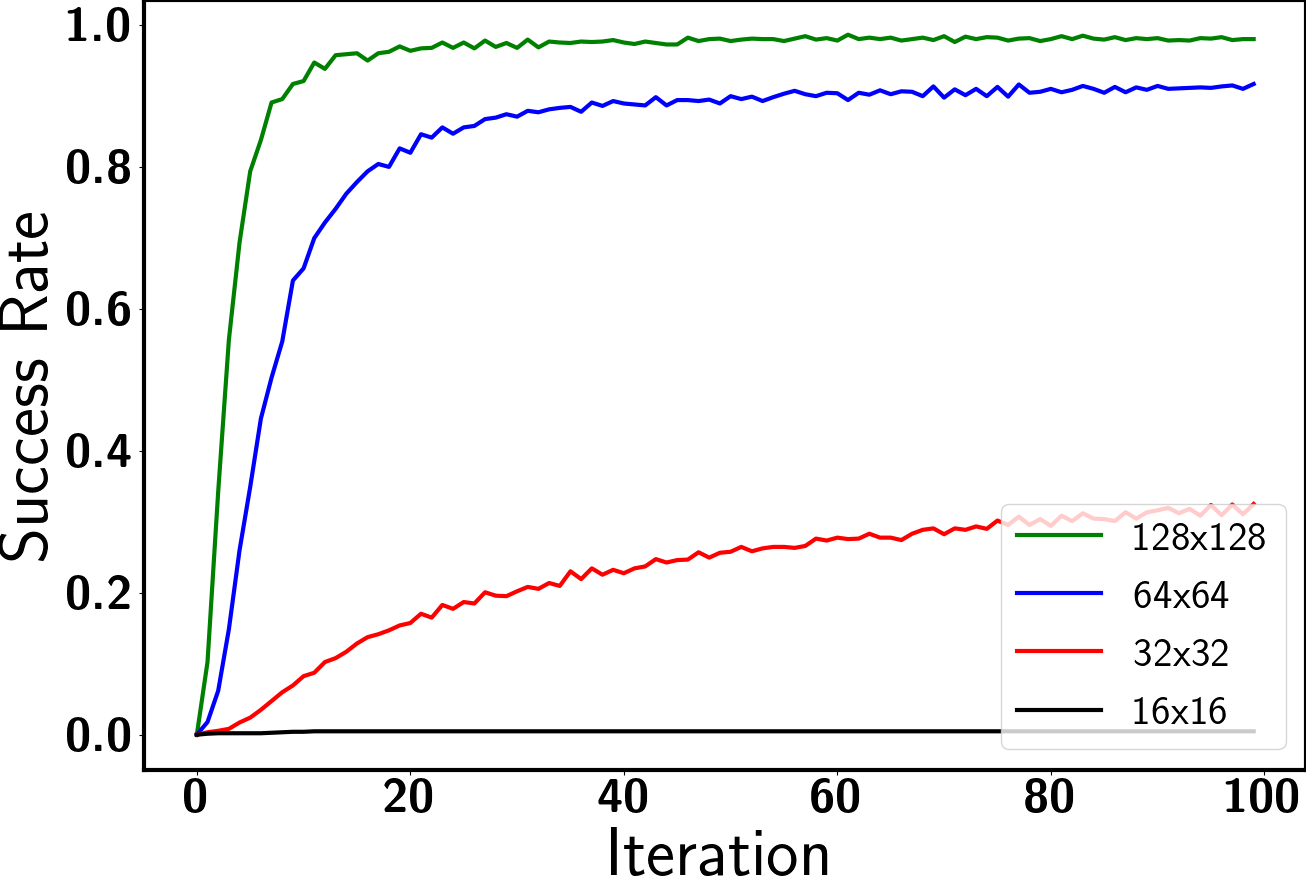}
    \caption{Different box sizes with $\xi=8, \alpha=2$.}
    \label{fig:box_suc}
\end{subfigure}
\caption{The success rate of the multi-untargeted attack with different $\xi$, $\alpha$, and box sizes.}
\label{fig.hyper_suc}
\end{figure*}

\begin{figure*}[t]
\centering
\begin{subfigure}[b]{0.32\textwidth}
    \centering
    \includegraphics[width=\textwidth]{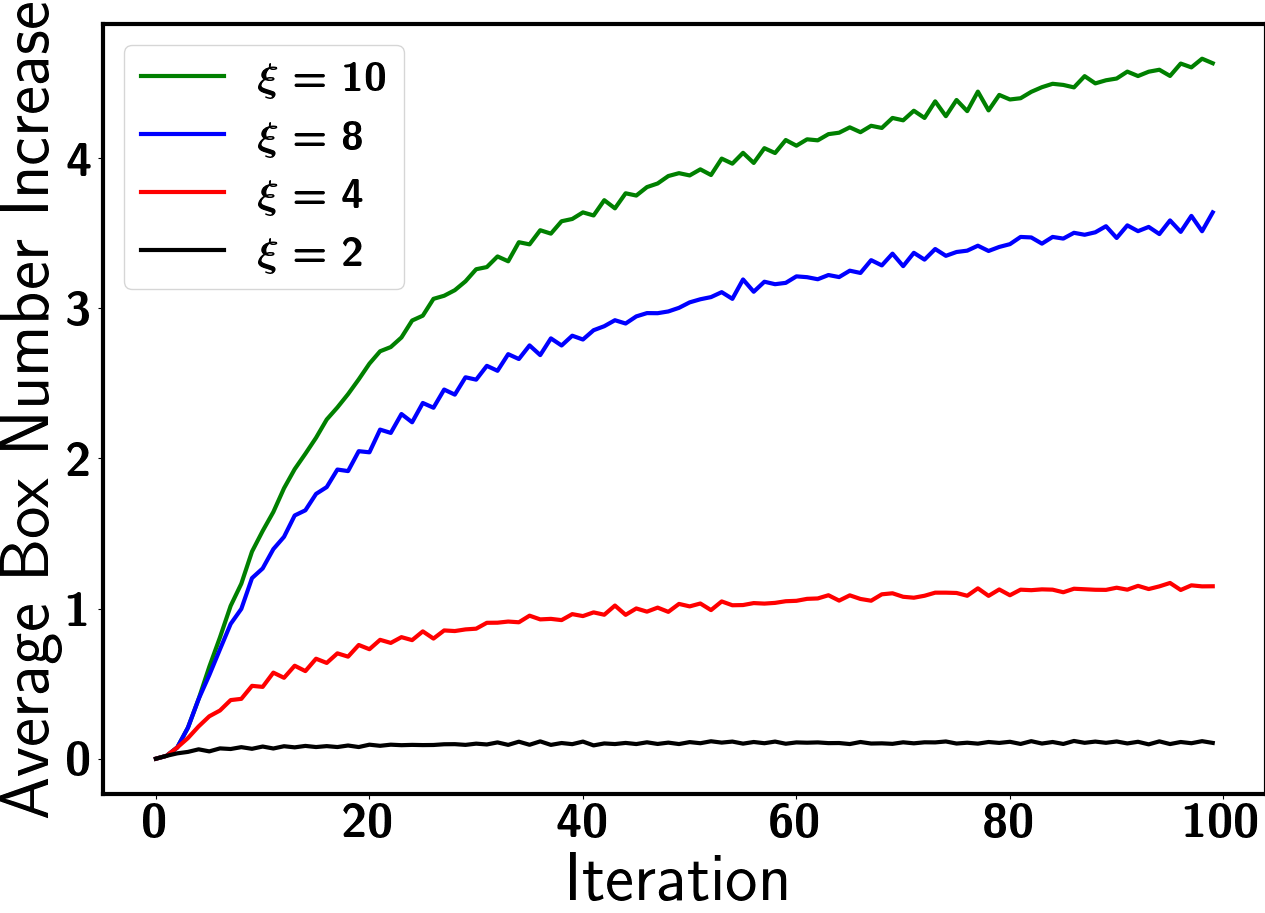}
    \caption{Different $\xi$ with $\alpha=2$, box size$=64$.}
    \label{fig:xi_box}
\end{subfigure}
\hfill
\begin{subfigure}[b]{0.32\textwidth}
    \centering
    \includegraphics[width=\textwidth]{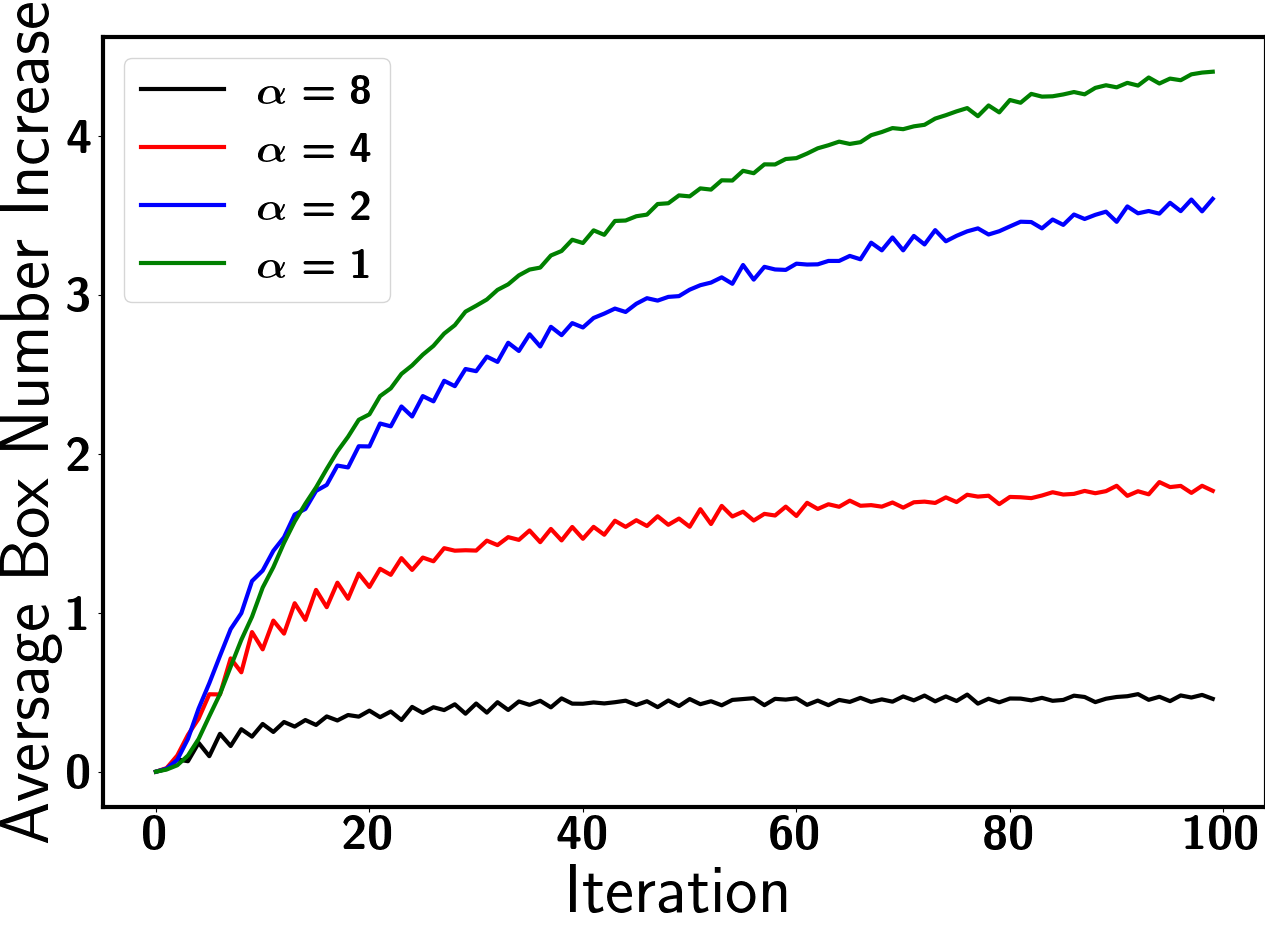}
    \caption{Different $\alpha$ with $\xi=8$, box size $=64$.}
    \label{fig:alpha_box}
\end{subfigure}
\hfill
\begin{subfigure}[b]{0.32\textwidth}
    \centering
    \includegraphics[width=\textwidth]{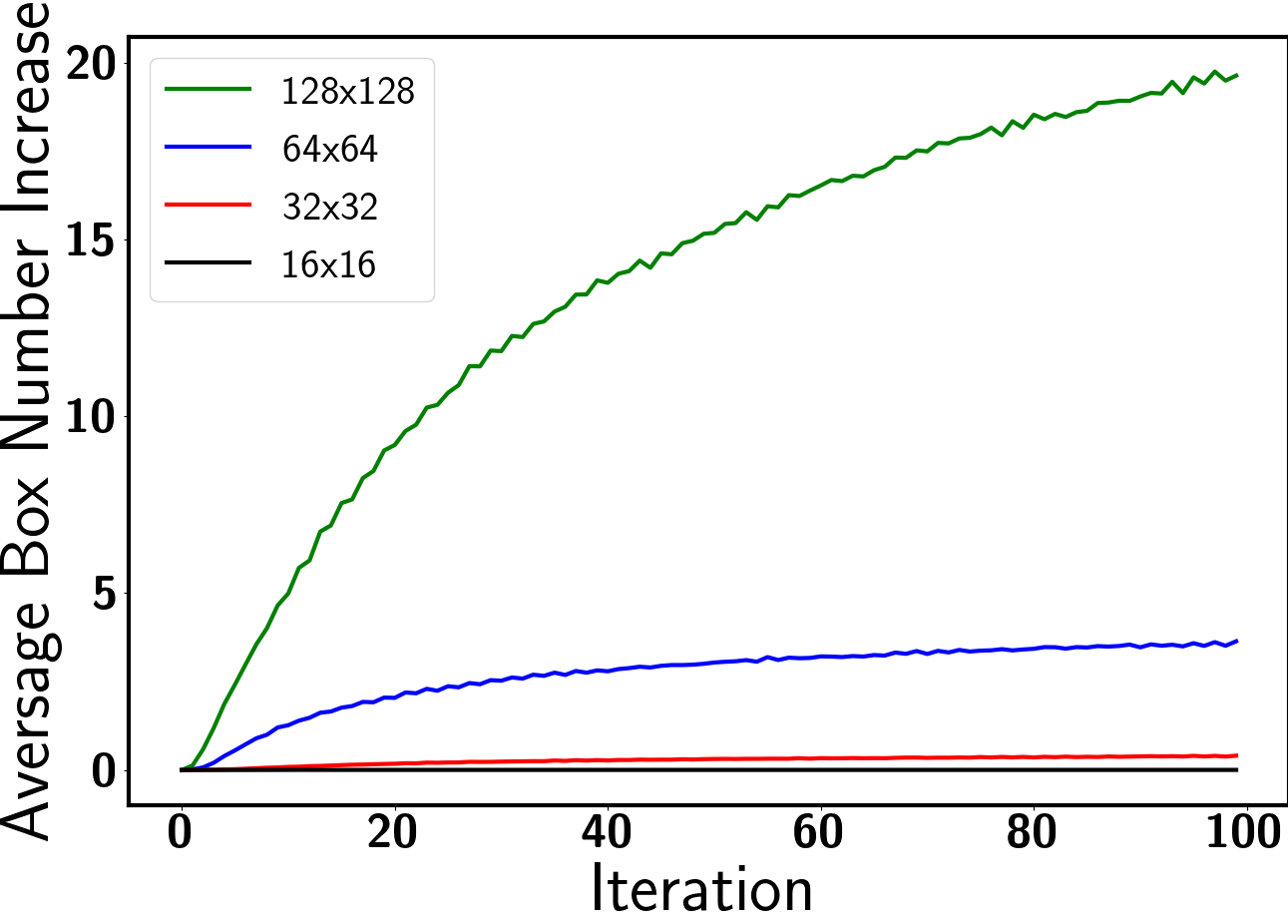}
    \caption{Different box sizes with $\xi=8, \alpha=2$.}
    \label{fig:box_box}
\end{subfigure}
\caption{The average box number increase of the multi-untargeted attack with different $\xi$, $\alpha$, and box sizes}
\label{fig.hyper_box}
\end{figure*}

\subsection{Evaluation Metrics}

Prior research have used the Mean Average Precision (mAP) \cite{map2018} to evaluate the accuracy of object detection models. However, the objective of our attack is to fabricate new bounding boxes, not to decrease the accuracy of existing objects. Therefore, the evaluation metrics need to reflect the efficiency of generating new bounding boxes in real time, something which mAP cannot do. In addition, the mAP evaluation metric requires access to the ground truth, that is, human-labeled bounding boxes. The online attack that fools the object detection system in real time does not have any such ground truth since it is impossible for humans to label the video stream in real time. Thus, we use the following two evaluation metrics that evaluate the efficiency of generating new objects and do not require ground truth labels.

\subsubsection{\textbf{Success Rate}}

The success rate measures the percentage of images we successfully fabricate at least one object within a limited number iterations. For a real-time online attack, we need to achieve a satisfying success rate within a limited number of iterations.

\subsubsection{\textbf{Mean Box Number Increase}}

We measure the number of new bounding boxes generated compared to the benign output. The more bounding boxes we generate, the stronger the perturbation is. However, to fabricate more objects we also need a larger number of iterations.

In general, we will attempt to find the minimum number of iterations required to achieve a satisfying success rate and fabricate enough bounding boxes. The PASCAL VOC-2012 validation set is used in the following experiments.

\subsection{Hyper-parameters}
\label{sec:hyper}

Three hyper-parameters are crucial for the attack: the attack strength $\xi$, the step size $\alpha$, and the box size. Since monochrome overlays can be generated in a less conspicuous way, we also investigated the effect of using gradients from different channels. In addition, we examined if the adversarial perturbation is sensitive to the aspect ratio.

\subsubsection{\textbf{The Attack Strength}}

The attack strength $\xi$ sets a boundary on the maximum perturbation added to each pixel. A larger $\xi$ makes the attack stronger but also results in a more conspicuous perturbation. In Fig. \ref{fig:xi_suc}, $\xi \in \{8, 10\}$ results in a much higher success rates than $\xi \in \{2, 4\}$ after 100 iterations. Though $\xi=8$ and $\xi=10$ share similar success rates, $\xi=10$ generates more bounding boxes than $\xi=8$ (see Fig. \ref{fig:xi_box}). For a real-time attack, we aim to generate at least one bounding box in 30 iterations, a requirement which is satisfied with both $\xi=8$ and $\xi=10$. $\xi=8$, in particular, finds a good balance between the attack strength and the impermeability of the perturbation.

\subsubsection{\textbf{The Step Size}}

The step size $\alpha$ controls how fast we update the perturbation in each iteration. A larger $\alpha$ would make it possible to fabricate bounding boxes in fewer iterations but make the update more unstable. In Fig. \ref{fig:alpha_suc}, both $\alpha=1$ and $\alpha=2$ achieve a success rate of more than 90\% after just 30 iterations, whereas $\alpha > 4$ produces a worse the success rate. In Fig. \ref{fig:alpha_box}, $\alpha=1$ eventually generates more bounding boxes than $\alpha=2$, but for a real-time online attack, it's more important to have a fast attack.


\begin{figure*}[t]
    \centering
    \begin{subfigure}[b]{0.32\textwidth}
        \includegraphics[width=\linewidth]{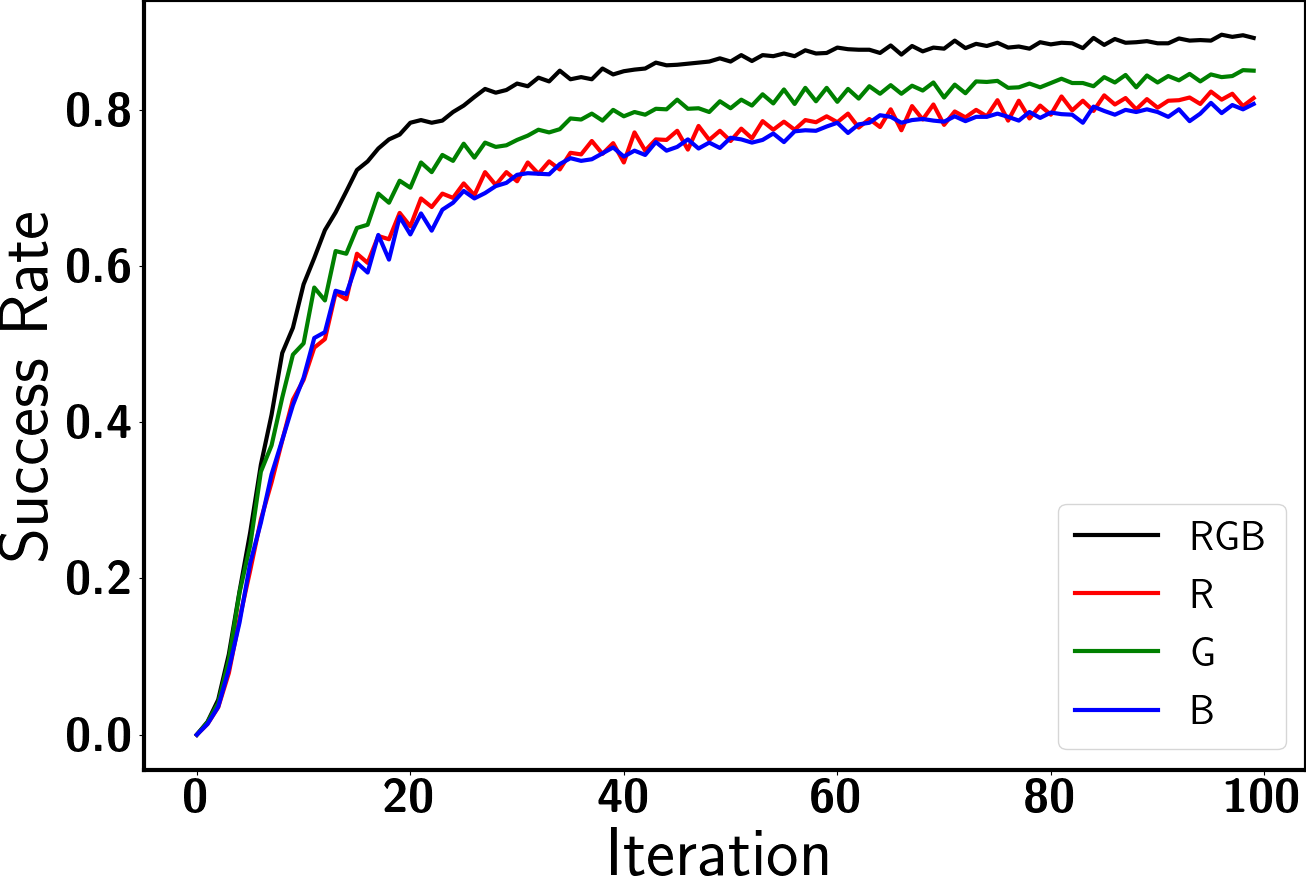}
        \caption{Different channels (Red, Green, Blue).}
         \label{fig:mono_suc}
    \end{subfigure}
    \begin{subfigure}[b]{0.64\textwidth}
        \includegraphics[width=1\linewidth]{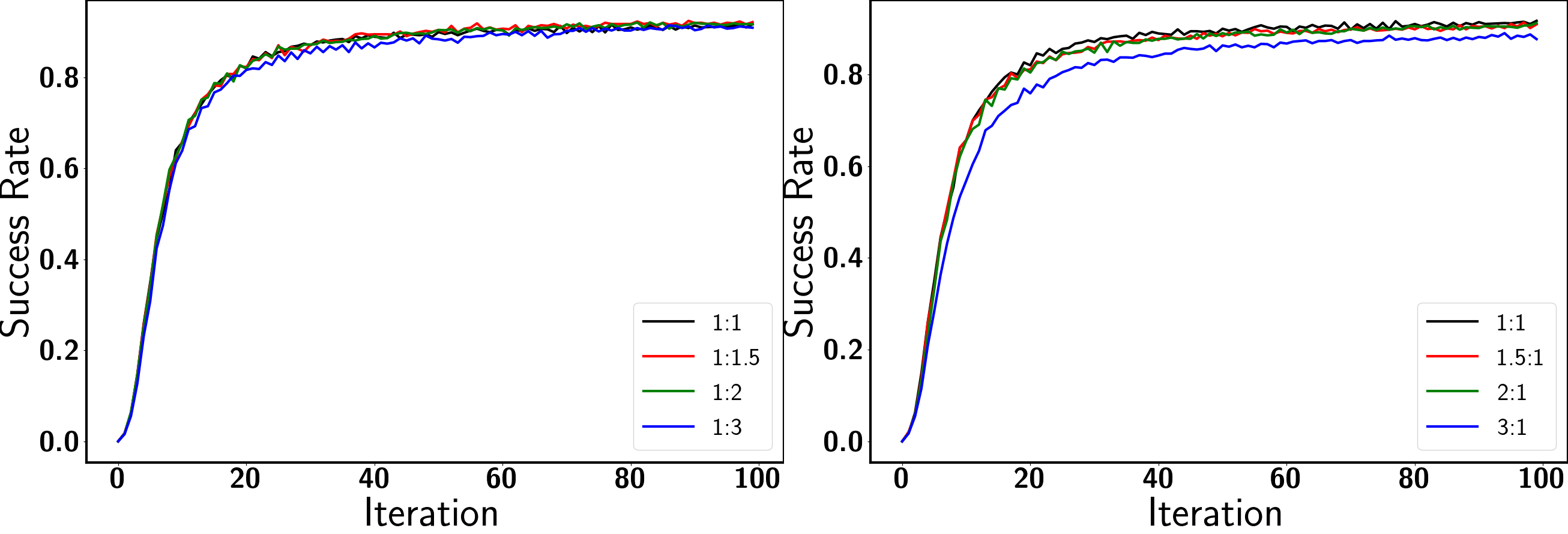}
        \caption{Different aspect ratios (left $1:n$, right: $n:1$).}
         \label{fig:aspect_suc}
    \end{subfigure}
    \caption{The success rate of the multi-untargeted attack with different channels and aspect ratios.}
\end{figure*}

\begin{figure*}[t]
    \centering
    \begin{subfigure}[b]{0.32\textwidth}
        \includegraphics[width=\linewidth]{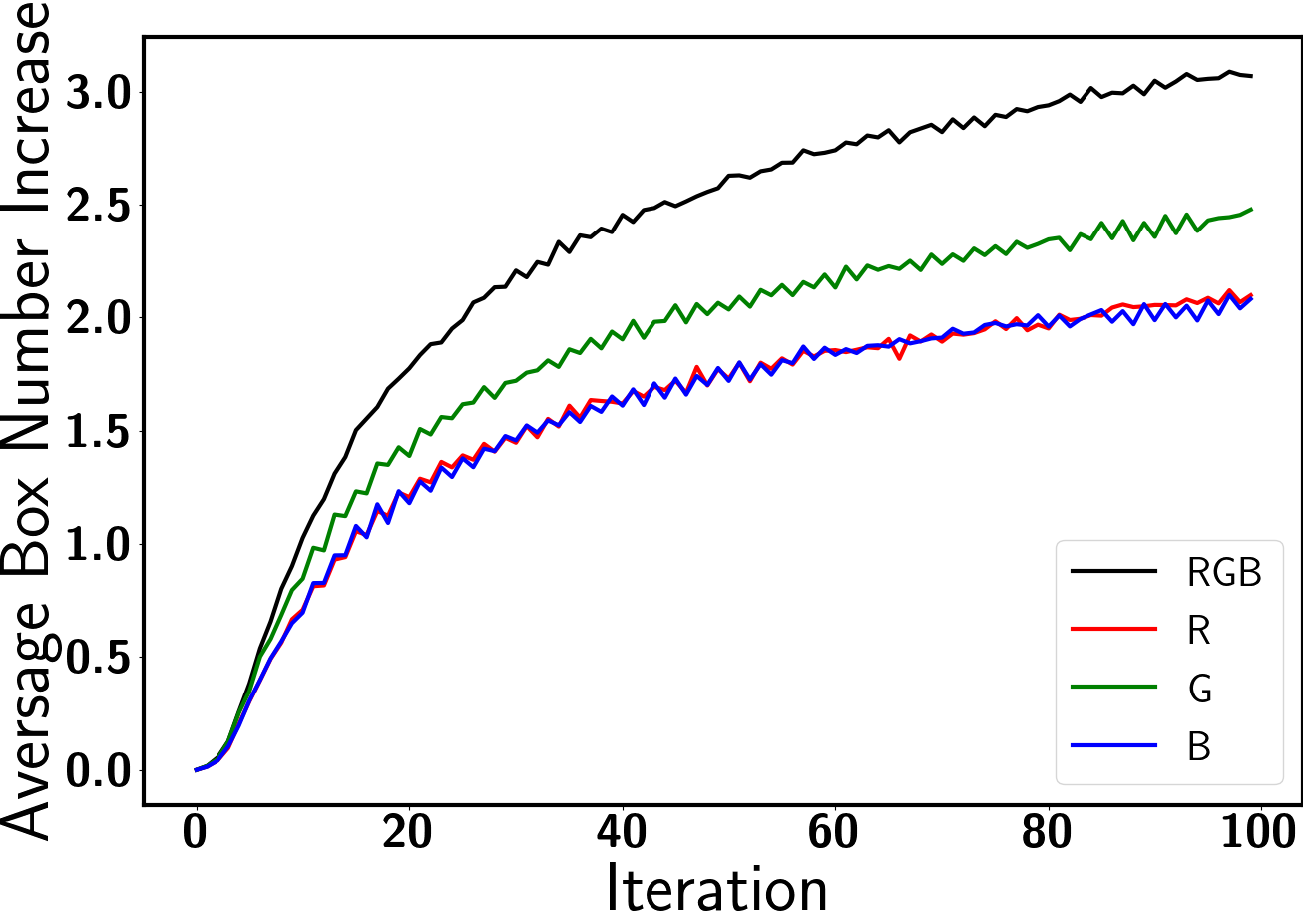}
        \caption{Different channels (Red, Green, Blue).}
         \label{fig:mono_box}
    \end{subfigure}
    \begin{subfigure}[b]{0.64\textwidth}
        \includegraphics[width=1\linewidth]{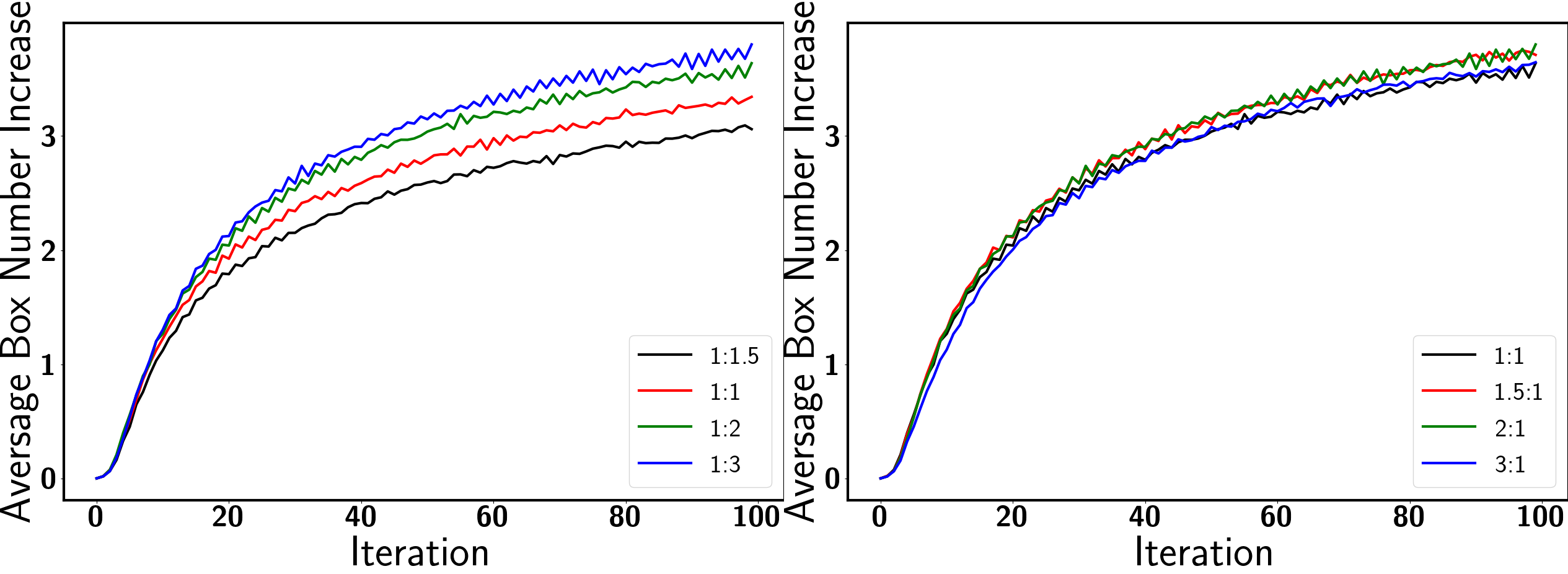}
        \caption{Different aspect ratios (left $1:n$, right: $n:1$).}
         \label{fig:aspect_box}
    \end{subfigure}

    \caption{The average box number increase of the multi-untargeted attack different channels and aspect ratios.}
\end{figure*}

\subsubsection{\textbf{The Box Size}}

The box size determines the total number of pixels we can perturb. The more pixels we can manipulate, the easier it is to deviate the model output. We need to find the minimum box size for our attack. In Fig. \ref{fig:box_suc}, the box sizes of both 64x64 and 128x128 achieve a success rate of 90\% after just 30 iterations, while the box size of 32x32 struggles to achieve successful attacks. Moreover, with box size smaller than 16x16 we are unable to attack the model with $\xi=8$ and $\alpha=2$. In addition, as expected, the larger the box size is, the more bounding boxes the attack can generate (see Fig. \ref{fig:box_box}). For a real-time online attack, we suggest the overlay box size to be at least 64x64.

\subsubsection{\textbf{The Monochrome Channel}}

For the attack that generates a polychrome overlay, we update each channel of the perturbation using the gradients of each channel. While for the monochrome overlay, we decide which color channel to prioritize. Interestingly, the experimental results (see Fig. \ref{fig:mono_suc} and Fig. \ref{fig:mono_box}) show that the object detection model is the most sensitive to the green channel. Using the average gradient of RGB channels, we achieve the highest success rate and generate the most number of boxes.

\subsubsection{\textbf{The Aspect Ratio}}

Our attack can generate adversarial overlays of arbitrary shapes. Thus, we can study if the object detection model is more susceptible to adversarial overlays of specific aspect ratios, e.g., wide (1:n) or long (n:1) boxes. For a fair comparison, overlay boxes of different aspect ratios have the same number of pixels perturbed. In Fig. \ref{fig:aspect_suc}, different aspect ratios do not cause observable dissimilarities in success rates. Although the aspect ratio $1:3$ generates more bounding boxes than $1:1.5$ after 100 iterations (see Fig. \ref{fig:aspect_box}), this attributes to a slightly more number of boxes perturbed ($1:3 \rightarrow 37 \times 111 = 4108$, $1:1.5 \rightarrow 52 \times 78 = 4056$). Thus, we cannot conclude that the aspect ratio has any discernible impact on the attack. As a result, we use $\xi=8$, $\alpha=2$, and a box size of 64x64 as default values for our attack, and the monochrome overlay uses the average of gradients from RGB channels.

\subsection{The Attack Performance}

We measured the performance of the attack on an NVIDIA RTX 2080Ti GPU. According to prior experimental results, we chose the hyper-parameters of $\xi=8$, $\alpha=2$, and used the box sizes of 64x64 and 128x128. Then, we tested the attack on the VOC 2012 validation set, which includes 5823 images in total (see Table \ref{tab:fix-it} and \ref{tab:fix-box}).


\clearpage

In Table \ref{tab:fix-it}, the attack achieved 24 FPS (1 iteration cost 41 ms). It should be noted that the performance of the attack also depends on model size. A larger model requires more computations to compute the gradient. Besides, the time cost does not grow as the box size increases (64 x 64 and 128 x 128) since we calculate the gradient of the adversarial loss functions over the entire input image, making it possible to generate multiple overlays of different shapes simultaneously without recalculating gradients.

\begin{table}[H]
    \centering
    \begin{tabular}{cccc}
    \hline
    Box Sizes & 1 iteration & 10 iterations & 20 iterations\\
    \hline
    \ 64x64    & 41 ms  & 410 ms & 780 ms \\
    \ 128x128  & 41 ms  & 410 ms & 781 ms \\
    \hline
    \end{tabular}
    \caption{The time cost of the attack with different numbers of iterations ($\alpha=2,\ \xi=8$).}
    \label{tab:fix-it}
\end{table}

\begin{table}[H]
    \centering
    \begin{tabular}{cccc}
    \hline
    Box Sizes & N=1 & N=3 & N=5\\
    \hline
    \ 64x64    & 7.47 it (306 ms)  & 12.03 it (493 ms) &  13.62 it (558 ms)) \\
    \ 128x128  & 4.40 it (180 ms)  & 7.64 it (313 ms) & 10.20 it (418 ms) \\
    \hline
    \end{tabular}
    \caption{The average number of iterations and time cost of generating N bounding boxes ($\alpha=2,\ \xi=8$).}
    \label{tab:fix-box}
\end{table}

As illustrated in Table \ref{tab:fix-box}, on average, the attack requires only 7 iterations within around 300 ms to generate one bounding box for each static image (offline attack). 

For an online attack, the attack can be even more efficient. It is unnecessary to re-generate the adversarial overlay from scratch for every input video frame since there is a high correlation between consecutive video frames and iterations \cite{ilyas2018prior}. As a result, we can save computations by reusing the overlay generated in the previous timestep. In the demo video, the attack achieved near real-time performance on an Intel i7-8665U CPU.

\section{Conclusions}

This paper has demonstrated that it is possible to attack an object detection system in real time. We generate human unperceivable adversarial overlays of arbitrary shapes to fabricate bounding boxes at desired locations. This attack could be a threat to the areas of traffic sign recognition and the autonomous driving field as a whole.

In the future, we plan to investigate the effect of the attack on modular autonomous driving systems that rely on object detection models to perceive the environment. In addition, we will explore how to detect adversarial attacks so that we can embrace deep learning models in safety-critical robotic applications in a safe way.











\bibliographystyle{IEEEtran}
\bibliography{IEEEabrv, mybibfile}

\end{document}